\pdfoutput=1

\documentclass[11pt]{article}
\usepackage[]{acl}

\usepackage{times}
\usepackage{latexsym}

\usepackage{CJKutf8}
\usepackage[T1]{fontenc}

\usepackage{amssymb}
\usepackage{graphicx}

\usepackage{soul}

\usepackage{multirow}
\usepackage{tabularx}
\usepackage{adjustbox}

\usepackage{bbding}
\usepackage{newverbs}
\usepackage[normalem]{ulem} 
\usepackage{makecell}
\usepackage{tabularray}

\usepackage{pgfplots} 
\pgfplotsset{width=8cm, height=3.75cm, compat=1.5}  
\usetikzlibrary{pgfplots.groupplots}

\usepackage{commath}

\usepackage{colortbl}
\usepackage[]{xcolor}
\definecolor{green}{RGB}{0.1,0.1,0.1}
\sethlcolor{lightgray}
\newcommand{\done}{\cellcolor{lightgray}}  

\usepackage{dblfloatfix}
\usepackage{float}
\usepackage{placeins}
\usepackage{MnSymbol}
\usepackage[utf8]{inputenc}
\usepackage{mathrsfs}
\usepackage{microtype}

\usepackage{amsmath}
\usepackage{float}
\usepackage{phaistos}
\title{Towards Accurate Translation via Semantically Appropriate Application of Lexical Constraints}

\author{Yujin Baek\thanks{\hspace{3pt} Equal Contribution}\hspace{5pt}$^{\Diamond}$, Koanho Lee\footnotemark[1]\hspace{5pt}$^{\Diamond}$, Dayeon Ki$^{\clubsuit}$ \\
    \textbf{Cheonbok Park$^{\heartsuit}$, Hyoung-Gyu Lee$^{\heartsuit}$  and Jaegul Choo$^{\Diamond}$} \\
  $^{\Diamond}$KAIST ,$^{\clubsuit}$Korea University,  $^{\heartsuit}$Papago, NAVER Corp.\\
  \texttt{\{yujinbaek, le5544, jchoo\}@kaist.ac.kr}\\
  \texttt{dayeonki@korea.ac.kr}, \texttt{\{cbok.park,hg.lee\}@navercorp.com} }

\begin{document}
\maketitle

\begin{table*}[h!]
\centering
\resizebox{2.0\columnwidth}{!}{%
\begin{tabular}{lllll}
\cline{4-5}
                                  &                                 &  & \multicolumn{2}{l}{\textbf{\done(Case A) Semantically Relevant Lexical Constraint}}                                                                                                                  \\ \cline{4-5} 
                                            &                                 &  & \multirow{2}{*}{\textbf{Source}}             & \multirow{2}{*}{\begin{CJK}{UTF8}{mj}\colorbox{lime}{\textbf{\textcolor{blue}{코로나}}} 이전 수준으로 경기가 완전히 회복하는 날이 올까요?\end{CJK}}                                                                                                \\
                                            &                                 &  &                                              &                                                                                                                                                  \\
                                            &                                 &  & \multirow{2}{*}{\textbf{Lexical Constraint}} & \multirow{2}{*}{\begin{CJK}{UTF8}{mj}\colorbox{lime}{\textbf{코로나}}\end{CJK} $\rightarrow$ \hl{\textbf{Covid-19}}
                                            \color{blue}{$$\CheckmarkBold$$ \texttt{{Automatically Retrieved from Bilingual Terminology}}}}                                                                                                      \\ \cline{1-2}
\multicolumn{2}{l}{\done$\;\;\;$\textbf{Bilingual Terminology}}                            &  &                                              &                                                                                                                                                  \\ \cline{1-2}
\multicolumn{1}{l|}{\textbf{Source Term}}   & \textbf{Target Term}            &  & \multirow{2}{*}{\textbf{Translation}}        & \multirow{2}{*}{Will the economy ever fully recover to before \hl{\textbf{Covid-19}} levels? \color{blue}{$$\CheckmarkBold$$}}                                                                  \\ \cline{1-2}
\multicolumn{1}{l|}{\multirow{2}{*}{\begin{CJK}{UTF8}{mj}선별진료소\end{CJK}}} & \multirow{2}{*}{Testing Center} &  &                                              &                                                                                                                                                  \\ \cline{4-5} 
\multicolumn{1}{l|}{}                       &                                 &  &                                              &                                                                                                                                                  \\ \cline{4-5} 
\multicolumn{1}{l|}{\multirow{2}{*}{\begin{CJK}{UTF8}{mj}\colorbox{lime}{\textbf{코로나}}\end{CJK}}}   & \multirow{2}{*}{\hl{\textbf{Covid-19}}}       &  & \multicolumn{2}{l}{\done\textbf{(Case B) Semantically Irrelevant Lexical Constraint}}                                                                                                                \\ \cline{4-5} 
\multicolumn{1}{l|}{}                       &                                 &  & \multirow{2}{*}{\textbf{Source}}             & \multirow{2}{*}{\begin{CJK}{UTF8}{mj}\colorbox{lime}{\textbf{\textcolor{red}{코로나}}} 엑스트라는 1998년 이후 미국에서 가장 많이 팔린 수입 음료이다.\end{CJK}}                                                                                       \\
\multicolumn{1}{l|}{\multirow{2}{*}{$\;\;\;\;\;\;\vdots$}}      & \multirow{2}{*}{$\;\;\;\;\;\;\vdots$}               &  &                                              &                                                                                                                                                  \\
\multicolumn{1}{l|}{}                       &                                 &  & \multirow{2}{*}{\textbf{Lexical constraint}} & \multirow{2}{*}{\begin{CJK}{UTF8}{mj}\colorbox{lime}{\textbf{코로나}}\end{CJK} $\rightarrow$ \hl{\textbf{Covid-19}} \color{red}{$$\XSolidBrush$$ \texttt{Automatically Retrieved from Bilingual Terminology}}}                                                                                                      \\ \cline{1-2}
                                            &                                 &  &                                              &                                                                                                                                                  \\
                                            &                                 &  & \multirow{3}{*}{\textbf{Translation}}        & \multirow{3}{*}{\begin{tabular}[c]{@{}l@{}}\textbf{\textcolor{red}{\sout{Covid-19}}} Extra has been the top-selling imported drink in the U.S. since 1998. \color{red}{$$\XSolidBrush$$}\\ \textbf{\textcolor{red}{Corona}}\end{tabular}} \\
                                            &                                 &  &                                              &                                                                                                                                                  \\
                                            &                                 &  &                                              &                                                                                                                                                  \\ \cline{4-5} 
\end{tabular}
}
\vspace{-2mm}
\caption{Automatically retrieved lexical constraint. }
\vspace{-4mm}
\label{tab:automatic-matching}
\end{table*}

\begin{abstract}

Lexically-constrained NMT (LNMT) aims to incorporate user-provided terminology into translations.
Despite its practical advantages, existing work has not evaluated LNMT models under challenging real-world conditions. 
In this paper, we focus on two important but understudied issues that lie in the current evaluation process of LNMT studies.
The model needs to cope with challenging lexical constraints that are ``homographs'' or ``unseen'' during training.
To this end, we first design a homograph disambiguation module to differentiate the meanings of homographs.
Moreover, we propose \texttt{PLUMCOT}, which integrates contextually rich information about unseen lexical constraints from pre-trained language models and strengthens a copy mechanism of the pointer network via direct supervision of a copying score.
We also release \texttt{HOLLY}, an evaluation benchmark for assessing the ability of a model to cope with ``homographic'' and ``unseen'' lexical constraints.
Experiments on \texttt{HOLLY} and the previous test setup show the effectiveness of our method. 
The effects of \texttt{PLUMCOT} are shown to be remarkable in ``unseen'' constraints.
Our dataset is available at \url{https://github.com/papago-lab/HOLLY-benchmark}.

%
%
%
%
%
\end{abstract}

\section{Introduction}
Lexically-constrained neural machine translation (LNMT) is a task that aims to incorporate pre-specified words or phrases into translations~(\citealp{GBS, Dinu, Song, Levenshtein, xu-carpuat-2021-editor, Cdalign, LeCA, Vecconst}, inter alia).
It plays a crucial role in a variety of real-world applications where it is required to translate pre-defined source terms into accurate target terms, such as domain adaptation leveraging domain-specific or user-provided terminology.
For example, as shown in Case A of Table~\ref{tab:automatic-matching}, an LNMT model successfully translates the source term (``\begin{CJK}{UTF8}{mj}코로나\end{CJK}'') into its corresponding target term (``Covid-19'') by adhering to a given lexical constraint (``\begin{CJK}{UTF8}{mj}코로나\end{CJK}” $\rightarrow$ ``Covid-19''). 

\textcolor{black}{Despite its practicality, previous studies on LNMT have not evaluated their performances under challenging real-world conditions. In this paper, we focus on two important but understudied issues that lie in the current evaluation process of the previous LNMT studies.}     
\paragraph{\emph{Semantics} of lexical constraints must be considered.} 
In previous work, at test time, lexical constraints are automatically identified from the source sentences by going through an \emph{automatic} string-matching process~\citep{Dinu, ailem2021lingua, LeCA}.
For example, in Case B of Table~\ref{tab:automatic-matching}, a source term (``\begin{CJK}{UTF8}{mj}코로나\end{CJK}'') in the bilingual terminology is present as a substring in the source sentence.
Accordingly, its corresponding target term (``Covid-19'') is automatically bound together as a lexical constraint (``\begin{CJK}{UTF8}{mj}코로나\end{CJK}” $\rightarrow$ ``Covid-19'') \textcolor{black}{without considering the semantics of the matched source term,\footnote{Here, \begin{CJK}{UTF8}{mj}코로나\end{CJK} in Case B indicates Corona, a brand of beer produced by a Mexican brewery.} which can lead to a serious mistranslation}.
This automatic string-matching cannot differentiate textually identical yet semantically different source terms.
\textcolor{black}{Thus, the more accurately the LNMT reflects the lexical constraint, the more pronounced the severity of the homographic issue is.}
\textcolor{black}{
To address this homograph issue, LNMT systems must be equipped to understand the semantics of identified lexical constraints and determine whether or not these constraints should be imposed.}

\paragraph{\emph{Unseen} lexical constraints need to be examined.}
\textcolor{black}{
One desideratum of LNMT systems is their robustness to handle ``unseen'' lexical constraints, thereby responding to random, potentially neologistic, or technical terms that users might bring up.
However, in previous studies, a significant portion of the lexical constraints is exposed during training. 
\citet{Vecconst} demonstrated the overlapped ratio of lexical constraints between the training and evaluation data (35.6\% on average).
Meanwhile,~\citet{zeng-etal-2022-neighbors} also raises the issue of the high frequency of lexical constraints for test sets appearing in the training data.
}

\textcolor{black}{
When lexical constraints are included in the training examples, we find that a well-optimized vanilla Transformer~\cite{Transformer} already satisfies lexical constraints by merely learning the alignment between the source and target terms co-occurring in the parallel training sentences.\footnote{We observe that the vanilla Transformer achieves a 66.67\% copy success rate.} 
This presents difficulties in identifying whether the presence of target terms in the output is attributed to the learned alignment, or the proposed components in previous studies.
Therefore, it is important to \textcolor{black}{control} lexical constraints not exposed during training to examine the model's ability to cope with ``unseen'' lexical constraints.
}

\textcolor{black}{
As a response, we present a \emph{test benchmark} for evaluating the LNMT models under these two critical issues. 
Our benchmark is specifically crafted not only to evaluate the performance of LNMT models but also to assess its ability to discern whether given lexical constraints are semantically appropriate or not.
To the best of our knowledge, we are the first to release a hand-curated high-quality test benchmark for LNMT.
Concurrently, we suggest a pipeline that allows researchers in LNMT communities to simulate realistic test conditions that consider the homograph issue and assign ``unseen'' lexical constraints.
}
 
\textcolor{black}{To this end, we propose a \emph{two-stage framework} to deal with these issues.
We first develop a homograph disambiguation module that determines whether LNMT models should apply a given lexical constraint by evaluating its semantic appropriateness.}
\textcolor{black}{Further, we propose an LNMT model that integrates provided lexical constraints more effectively by learning when and how to apply these lexical constraints.} \textcolor{black}{Our contributions are summarized as follows:
\begin{itemize}
    \setlength\itemsep{0em}
    \item  
    We formulate the task of semantically appropriate application of lexical constraints and release a high-quality test benchmark to encourage LNMT researchers to consider real-world test conditions.
    \item  We propose a novel homograph disambiguation module to detect semantically inappropriate lexical constraints. 
    \item \textcolor{black}{We present an LNMT model which shows the best translation quality and copy success rate in unseen lexical constraints. }
\end{itemize}
}

\begin{table*}[t!]
   \resizebox{2\columnwidth}{!}{
  \begin{tabular}{c|lll|l}
\hline\hline
\textbf{Source Term}& \multicolumn{3}{l|}{\textbf{Test Example}}                                                      & \textbf{Lexical Constraint}                  \\ \hline
\multirow{12}{*}{\begin{CJK}{UTF8}{mj}\textbf{양수}\end{CJK}} 
& \multirow{3}{*}{(a)} & Src. & \begin{CJK}{UTF8}{mj}\colorbox{lime}{\textcolor{blue}{\textbf{양수}}} 파열로 출산이 임박한 산모가 공군의 도움으로 건강한 아이를 출산했다.\end{CJK} & \multirow{3}{*}{\begin{CJK}{UTF8}{mj}양수\end{CJK} $\rightarrow$ amniotic fluid  \color{blue}{$$\CheckmarkBold$$}} \\
      & &  Ref. & \makecell[l]{A pregnant woman on the verge of labor due to \hl{amniotic fluid} breaking gave birth to a \\ healthy child thanks to the help of the airforce.}  &                              \\ \cline{2-5} 
& \multirow{2}{*}{(b)} & Src. & \begin{CJK}{UTF8}{mj}평소 다니던 산부인과 의사 선생님이 \colorbox{lime}{\textcolor{blue}{\textbf{양수}}} 검사를 권해서 하고 왔습니다.\end{CJK} & \multirow{2}{*}{\begin{CJK}{UTF8}{mj}양수\end{CJK} $\rightarrow$ amniotic fluid \color{blue}{$$\CheckmarkBold$$}}            \\ 
      & &  Ref. & As my regular gynecologist recommended an \hl{amniotic fluid} test, I took the test and came back.  &                            \\ \cline{2-5} 
& \multirow{3}{*}{(c)} & Src. & \makecell[l]{\begin{CJK}{UTF8}{mj}수학에서는 \colorbox{lime}{\textcolor{red}{\textbf{양수}}}를 나타낼 때는 `＋' 기호를 생략해도 되지만 음수를 나타낼 때에는 반드시\end{CJK}\\ \begin{CJK}{UTF8}{mj} `－' 기호를 숫자 앞에 붙여야 한다.\end{CJK}                                                }                   & \multirow{3}{*}{\begin{CJK}{UTF8}{mj}\sout{양수 $\rightarrow$ amniotic fluid}\end{CJK} \color{red}{$$\XSolidBrush$$}}            \\
      & &  Ref. & \makecell[l]{In mathematics, while you can omit the `+' symbol when indicating \hl{positive}, you must mark \\ the `-' one before numbers when meaning negative.}   &                              \\ \cline{2-5} 
& \multirow{3}{*}{(d)} & Src. & \begin{CJK}{UTF8}{mj}상장 법인 대주주들의 주식 양도 \colorbox{lime}{\textcolor{red}{\textbf{양수}}}가 최근 들어 활발한 것으로 나타났다.\end{CJK}                                                                   & \multirow{3}{*}{\begin{CJK}{UTF8}{mj}\sout{양수 $\rightarrow$ amniotic fluid}\end{CJK} \color{red}{$$\XSolidBrush$$} }            \\
       &      & Ref. & \makecell[l]{It turns out that recently, major shareholders of the listed corporates have active handovers\\ and \hl{takeovers}.} &                              \\ \hline\hline
\end{tabular}}
\caption{Test examples are bound together with a lexical constraint. (a) and (b) are positive examples. (c) and (d) are negative examples. The source term can be pronounced as ``yang-soo''. Four examples are assigned to each lexical constraint.}
\label{table: evaluation data} 
\end{table*}

\begin{table*}[t!]
\centering
  \resizebox{2\columnwidth}{!}{
\begin{tabular}{l|ll}
\hline\hline
$\;\;$\textbf{Lexical Constraint }                               & \multicolumn{2}{l}{\textbf{Positive Reference}}                                                                     \\ \hline
\multirow{4}{*}{$\;\;$\begin{CJK}{UTF8}{mj}\textbf{양수}\end{CJK} $\rightarrow$ amniotic fluid \color{blue}{$$\CheckmarkBold$$}}$\;\;\;\;\;\;$  & Src. & \begin{CJK}{UTF8}{mj}출산 예정일보다 일찍 \colorbox{lime}{\textcolor{blue}{\textbf{양수}}}가 터지는 경우가 있다.\end{CJK}                                                                         \\
                                                 & Ref. & (There are cases where the \hl{amniotic fluid} bursts sooner than the expected date of birth.)           \\ \cline{2-3} 
                                                 & Src. & \begin{CJK}{UTF8}{mj}태아의 염색체 이상 여부를 알아보기 위해 \colorbox{lime}{\textcolor{blue}{\textbf{양수}}}를 검사했다.\end{CJK}                                                                    \\
                                                 & Ref. & (The \hl{amniotic fluid} was tested to find if there were any abnormalities with the fetal chromosomes.) \\ \hline
\end{tabular}}
\vspace{-1mm}
\caption{Positive References. Homograph (\begin{CJK}{UTF8}{mj}\colorbox{lime}{\textcolor{blue}{\textbf{양수}}}\end{CJK}) in positive references means the \hl{amniotic fluid}.}
\vspace{-3mm}
\label{table: positive references}
\end{table*}

\section{\texttt{HOLLY} Benchmark} 
Here, we introduce \texttt{HOLLY} (\underline{\textbf{ho}}mograph disambiguation evaluation for lexica\underline{\textbf{lly}} constrained NMT), a novel benchmark for evaluating LNMT systems in two circumstances; either the assigned lexical constraints are semantically appropriate or not, as illustrated in Table~\ref{table: evaluation data}.
The entire test data includes 600 test examples on 150 Korean $\rightarrow$ English lexical constraints.

\subsection{Test Examples}\label{sec:pos-neg-examples}

Each test example consists of three main elements, as presented in Table~\ref{table: evaluation data}: (1) a lexical constraint (\begin{CJK}{UTF8}{mj}양수\end{CJK} $\rightarrow$ amniotic fluid), (2) a source sentence containing the source term (\begin{CJK}{UTF8}{mj}\colorbox{lime}{양수}\end{CJK}) of the lexical constraint, and (3) its reference translation.\footnote{\color{black}{We outsourced the translation process to a professional translation company, and each translation was manually reviewed.}}

\textcolor{black}{
While the source term is a homograph with multiple meanings, one of them is chosen to serve as its lexical constraint.\footnote{Out of multiple different meanings of a homograph, we select the least frequent one as its lexical constraint. We describe the data construction details in Appendix~\ref{appendix: Source of Data}.}}
\textcolor{black}{Then, based on the meaning of the source term in the source sentence, each test example is classified into one of two groups:}
\begin{itemize}
    \setlength\itemsep{0em}
    \item  \textbf{\textit{Positive Example}} where the source term in its source sentence is semantically aligned to the lexical constraint (See test examples (a) and (b) in Table~\ref{table: evaluation data}). \textcolor{black}{For positive test examples, we expect lexical constraints should always be applied.}
    \item \textbf{\textit{Negative Example}} where the given lexical constraint is semantically improper to impose (See test examples (c) and (d) in Table~\ref{table: evaluation data}). \textcolor{black}{Negative test examples allow us to evaluate how LNMT models respond to inappropriate lexical constraints.}
\end{itemize}
\subsection{Positive References}
As seen in Table~\ref{table: positive references}, we provide two auxiliary source-side example sentences demonstrating the specific use of the source term of its lexical constraint, assuming that the meaning can be differentiated by the context used in the sentences rather than the terminology itself.
Hereafter, we name these example sentences as \textit{positive references}.






\textcolor{black}{\section{Methodology}
Our methodology for semantically appropriate application of lexical constraints consists of two stages.
Initially, we propose a \emph{homograph disambiguation module} that can differentiate the semantics of lexical constraints.
This module determines whether LNMT models should incorporate a lexical constraint or not.
Subsequently, LNMT models, \texttt{PLUMCOT} in our case, perform the translation, either with or without the given lexical constraints.
}

\subsection{Homograph Disambiguation}
Given a few example sentences demonstrating how to specifically use a word, humans can infer the proper meaning.
Likewise, our conjecture is that we can fulfill the homograph disambiguation task by leveraging these inter-sentential relationships.

\subsubsection{Task Specification} 
\label{Sec: Task Specification}
\begin{figure}
\centering  \includegraphics[width=0.9\columnwidth,height=0.7\columnwidth]{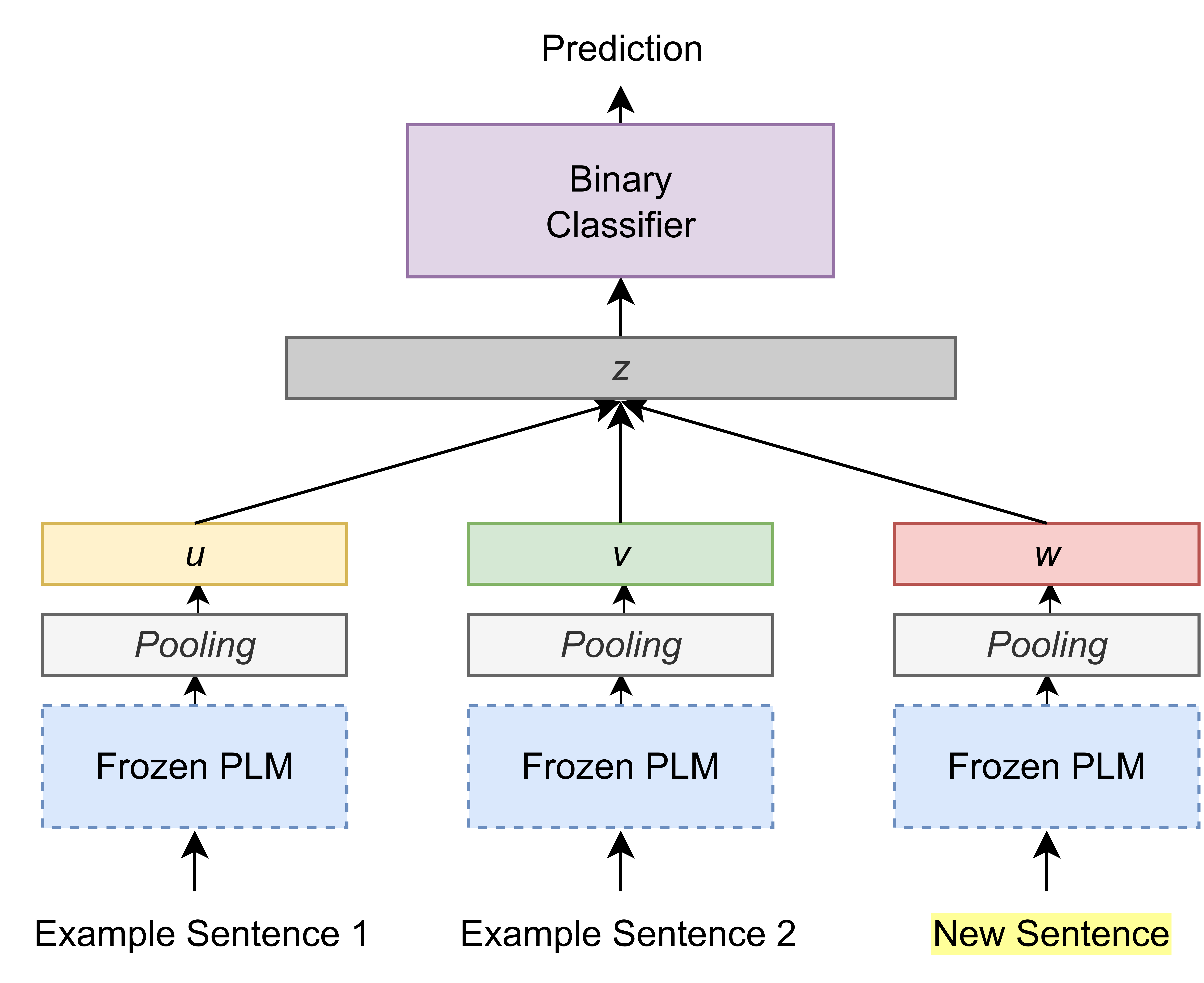}
  \vspace{-3mm}
  \caption{Structure of the homograph disambiguation module.}
  \vspace{-4mm}
  \label{fig:homograph disambiguation module}
\end{figure}

Given $n$ example sentences illustrating one specific meaning of a homograph, our \textit{homograph disambiguation module} aims to determine whether the same word in a newly given sentence, denoted as `New Sentence' in Fig.~\ref{fig:homograph disambiguation module}, carries the same meaning (\textbf{label: 1}) or not (\textbf{label: 0}).
We conducted experiments with two example sentences (i.e., $n =2$),\footnote{We experiment with $n=1, 2,\text{and } 3$. The effect of varying the number of example sentences is analyzed in Appendix~\ref{appendix: Number of example sentences}.} and the corresponding model architecture is described in Section~\ref{sec:hdm_architecture}. 

\subsubsection{Model Architecture}\label{sec:hdm_architecture}
\paragraph{Input Representations}
As illustrated in Fig.~\ref{fig:homograph disambiguation module}, sentence embeddings of example sentences and the new sentence are individually obtained from the PLM and fed into the classifier.
Embedding vectors for all the sentences are extracted from the averaged hidden representations of the last $K$ layers of frozen PLM.\footnote{We utilize the last 16 layers of \textit{klue/roberta-large}, a RoBERTa-based PLM trained on Korean corpus. See \url{https://huggingface.co/klue/roberta-large} for details.}
\textcolor{black}{
Here, the embedding vector is obtained by the average of hidden representations for the tokens that make up a homograph within the sentence.
}
We denote this averaging operation as \emph{Pooling} in Fig.~\ref{fig:homograph disambiguation module}.

\begin{table*}[t]
\centering
   \resizebox{2.0\columnwidth}{!}{
\begin{tabular}{ll}
\hline
\multicolumn{2}{l}{$\;\;$\textbf{(Lexical Constraint)} \begin{CJK}{UTF8}{mj}\colorbox{lime}{\textbf{\textcolor{blue}{코로나}}}\end{CJK} $\rightarrow$ \hl{\textbf{Covid-19}} \color{blue}{$$\CheckmarkBold$$\texttt{\textbf{Approved} by homograph disambiguation module}}}       \\ \hline\hline
\multicolumn{1}{l|}{\multirow{2}{*}{$\;\;$\textbf{Source Sentence}$\;\;$}}          & \multirow{2}{*}{\makecell{\begin{CJK}{UTF8}{mj}\colorbox{lime}{\textbf{\textcolor{blue}{코로나}}} 이전 수준으로 경기가 완전히 회복하는 날이 올까요?\end{CJK}\\ (Will the economy ever fully recover to before \hl{\textbf{Covid-19}} levels?)}}                              \\
\multicolumn{1}{l|}{}                                          &                                                      \\ \hline
\multicolumn{1}{l|}  {$\;\;$\textbf{Modified Source Sentence}$\;\;\;\;\;\;$} & \begin{CJK}{UTF8}{mj}\colorbox{lime}{\textbf{\textcolor{blue}{코로나}}} 이전 수준으로 경기가 완전히 회복하는 날이 올까요?\end{CJK}$\;$<sep>$\;$\hl{\textbf{Covid-19}} \\ \hline
\end{tabular}}

$\;$\newline

\centering
   \resizebox{2.0\columnwidth}{!}{%
\begin{tabular}{ll}
\hline
\multicolumn{2}{l}{$\;\;$\textbf{(Lexical Constraint)} \begin{CJK}{UTF8}{mj}\colorbox{lime}{\textbf{\textcolor{red}{코로나}}}\end{CJK} $\rightarrow$ \hl{\textbf{Covid-19}} \color{red}{$$\XSolidBrush$$ \texttt{\textbf{NOT approved} by homograph disambiguation module}}}                                                                                 \\ \hline\hline
\multicolumn{1}{l|}{\multirow{2}{*}{$\;\;$\textbf{Source Sentence}$\;\;$}}          & \multirow{2}{*}{\makecell{\begin{CJK}{UTF8}{mj}\colorbox{lime}{\textbf{\textcolor{red}{코로나}}} 엑스트라는 1998년 이후 미국에서 가장 많이 팔린 수입 음료이다.\end{CJK}\\ (Corona Extra has been the top-selling imported drink in the U.S. since 1998.)}}                              \\
\multicolumn{1}{l|}{}                                          &                                                      \\ \hline
\multicolumn{1}{l|}  {$\;\;$\textbf{Modified Source Sentence}$\;\;\;\;\;\;$} & \begin{CJK}{UTF8}{mj}\colorbox{lime}{\textbf{\textcolor{red}{코로나}}} 엑스트라는 1998년 이후 미국에서 가장 많이 팔린 수입 음료이다.\end{CJK} \\ \hline
\end{tabular}}
\caption{Input modification. Expected target terms are appended to the end of the source sentence. The source term can be pronounced as ``co-ro-na''.}
\vspace{-3mm}
\label{tab:input-data}
\end{table*}
\paragraph{Binary Classifier}


\textcolor{black}{Similar to Sentence-BERT \cite{reimers2019sentence}, we use the concatenation ($z \in \mathbb{R}^{6m+3}$) of the following as an input to the classifier:}
\begin{itemize}
    \setlength\itemsep{0em}
    \item \textcolor{black}{Contextualized representation of a homograph ($u, v, w \in \mathbb{R}^{m}$),}
    \item element-wise difference for each pair ($\abs{u-v}, \abs{v-w}, \abs{u-w} \in \mathbb{R}^{m}$,)
    \item pair-wise cosine similarity scores ($sim(u,v), \;sim(v,w), \;sim(u,w) \in \mathbb{R}$), 
\end{itemize}

where $m$ is the dimension of the embeddings and $sim(\cdot, \cdot)$ denotes the cosine similarity function. Our prediction $o \in [0,1]$ for a `New Sentence' is calculated as
\begin{equation}
o =  \sigma(\max(0, {zW_{r}+b_{r}})W + b),
\end{equation}
where $W_{r} \in \mathbb{R}^{(6m+3)\times m}$ and $b_{r}$ are the weight matrix and bias vector of an intermediate layer, respectively.
$W \in \mathbb{R}^{m\times 1}$ and $b$ are the weight matrix and bias vector for the final prediction layer followed by $\sigma(\cdot)$, which represents the sigmoid function. 

\subsection{\texttt{PLUMCOT}}
\label{sec: PLUMCOT}
\textcolor{black}{In this subsection, we introduce our LNMT model, \texttt{PLUMCOT}, which stands for leveraging \underline{p}re-trained \underline{l}ang\underline{u}age \underline{m}odel with direct supervision on a \underline{co}pying score for LNM\underline{T}, and its detailed implementation.}
To better incorporate target terms into the translations, \texttt{PLUMCOT} combines LeCA~\cite{LeCA} with PLM and strengthens a pointer network with supervised learning of the copying score.

\subsubsection{Problem Statement}
\paragraph{Lexically-constrained NMT}

Suppose $X = (x_1, x_2, \cdots, x_{|X|})$ as a source sentence and $Y = ( y_1, y_2, \cdots, y_{|Y|} )$ as a target sentence. 
Given the constraints $C = (C_1, C_2, \cdots, C_n)$ where each constraint $C_i = (C_{i,S}, C_{i,T})$ consists of the source term $C_{i,S}$ and corresponding target term $C_{i,T}$, LNMT aims to incorporate $C_{1:n,T}$ into its generation. 
The conditional probability of LNMT can be defined as
\begin{equation}
\label{eqn:conditional prob}
\begin{aligned}
p(Y|\;X, C;\theta) = 
\displaystyle \prod_{t=1}^{|Y|} p(y_t|y_{<t}, X, C ; \: \theta).
\end{aligned}
\end{equation}

\paragraph{Input Data}\label{sec: Input data}

As in~\citet{LeCA}, we modify $X$ as $\hat{X} = ( X, \text{<sep>}, C_{1, T}, \cdots, \text{<sep>},  C_{n, T})$ by appending $\text{<sep>}$ tokens followed by target terms, as illustrated in Table~\ref{tab:input-data}.\footnote{In our training, we randomly sample target terms from the target sentence. Please refer to Appendix~\ref{appendix: input data augmentation} for details.}
If there are no lexical constraints, a source sentence remains the same, i.e., $\hat{X} = X$.\footnote{\textcolor{black}{At test time, we append target terms only when lexical constraints are determined to be used by the homograph disambiguation module (as indicated in Table~\ref{tab:input-data}).}}
Combining a source sentence with target terms leads to the modification of Eq.~\eqref{eqn:conditional prob} as the following:
\begin{equation}
p(Y|\;X, C;\theta) = 
\displaystyle \prod_{t=1}^{|Y|} p(y_t|y_{<t}, \hat{X} ; \: \theta).
\end{equation}

\subsubsection{Integration of PLM}
\label{body: PLM}
As PLM such as BERT~\cite{Bert} is trained on large amounts of unlabeled data, leveraging PLM for LNMT can provide rich contextualized information of $X$, even in controlled unseen lexical constraint scenarios.


\textcolor{black}{We first feed the source sentence $X$ to a frozen PLM to obtain a representation $B$ of a source sentence, where $B$ is the output of the last layer of the PLM.
Conversely, our NMT model based on~\citet{Transformer} receives a modified source sentence $\hat{X}$ as input.} 

\textcolor{black}{Let $L$ denote the number of encoder and decoder layers of NMT, $H^{l}$ be the output of the encoder of NMT at the $l$-th layer, and $h_{t}^{l}$ denote the $t$-th element of $H^{l}$.}
\textcolor{black}{For each layer $l \in [1, L]$, we employ multi-head attention with the output of PLM as in~\citet{Bert-fused}, denoted as $ \text{MHA}_{\text{B}}$.
\textcolor{black}{This maps the output of the NMT encoder at $l-1\text{th}$ layer into queries and output of PLM, $B$, into keys and values.\footnote{Please refer to Appendix~\ref{Appendix: MHA} for more details.} }
The output of the $t$-th element of the NMT encoder at the $l$-th layer is given by}
\begin{equation}\label{eqn: Body MHA}
\begin{aligned}
    \tilde{h}_{t}^{l} &= 
    \frac{1}{2} \big( \text{MHA} (h_{t}^{l-1},  H^{l-1},  H^{l-1})  \\
    &+ \text{MHA}_{\text{B}} (h_{t}^{l-1},  B,  B ) \big) +  h_{t}^{l-1},
    \\
    h_{t}^{l} &= \text{LN}\big(\text{FFN}(\text{LN}(\tilde{h}_{t}^{l})) + \tilde{h}_{t}^{l} \big),
\end{aligned}
\end{equation}
where $\text{LN}(\cdot)$ denotes Layer normalization in ~\citet{ba2016layer} and $\text{MHA}$ and $\text{FFN}(\cdot)$ are the multi-head attention and feed-forward network, respectively.

Similar to the encoder, multi-head attention with PLM is introduced for each decoder layer.\footnote{Please refer to Appendix~\ref{Appendix: Integration of PLM in Decoder} for more details.} 
Combined with Section~\ref{sec:supervision on a copying score}, a highly contextualized representation is given to the pointer network.



\subsubsection{Supervision on a Copying Score}
\label{sec:supervision on a copying score}
\paragraph{Pointer Network}
To copy target terms from $\hat{X}$, we introduce a pointer network~\cite{gu2016incorporating} as in~\citet{Song, LeCA}. \textcolor{black}{For each time step, a pointer network takes in the output of the encoder and outputs a copying score $g_t^\text{copy} \in [0, 1]$, which controls how much to copy. 
The output probability of the target word $y_t$ can be calculated as}
\begin{equation}
\label{eqn:output probability}
   p(y_t|y_{<t}, \hat{X} ; \: \theta)  = (1 - g_t^\text{copy}) \times p_{t}^\text{word} + g_t^\text{copy} \times p_{t}^\text{copy},
\end{equation}
where $p_{t}^\text{copy}$ is a probability of copying, and $p_{t}^\text{word}$ is a probability of the target word $y_t$ in the vocabulary.\footnote{Please refer to Appendix~\ref{appendix: Pointer Network} for more details.}

\paragraph{Copying Score}
As implied by Eq.~\eqref{eqn:output probability}, inaccurately predicted $g_t^\text{copy}$ results in the failure of copying target terms. 
\textcolor{black}{
However, in previous research on LNMT, the importance of a copying score was relatively understated.
Despite the high probability of copying $p_{t}^\text{copy}$, \textcolor{black}{an incorrect copying score can even lower the output probability of the target terms.}
Therefore, we propose a novel supervised learning of the copying score $g_t^\text{copy}$ to obtain a more accurate value.
}

\textcolor{black}{Our supervision of the copying score strengthens the copy mechanism of the pointer network by allowing the model to learn exactly when to copy.}
Since target terms are in the source sentence, we can determine which words should be copied from the source sentence. 
For example, when translating a source sentence in Table~\ref{tab:input-data}, the appended target term, \hl{Covid-19}, must be copied. 
Thus, the copying score $ g_t^\text{copy}$ of the target term \hl{Covid-19} should be higher, and 
\textcolor{black}{
$ g_t^\text{copy}$ should be lower for the remaining words in the target sentence.}
Our training objective can be defined as
\begin{equation}
\begin{aligned}
 L(\theta) &= - \displaystyle\sum_{t=1}^{|Y|}\;\log p(y_t|y_{<t}, \hat{X} ; \; \theta) -  \lambda J(\theta), \\
J(\theta) &=
\alpha  \displaystyle\sum_{t\notin C_{{1:n}, T}} (1 - g_{t}) \times \log \; ( 1- g_{t}^\text{copy}) \\
&+ \beta \displaystyle \sum_{t \in C_{{1:n}}, T} g_{t} \times \log \;  g_{t}^\text{copy},
\end{aligned} \label{eqn: gate classification}
\end{equation}
where a gold copying score $g_{t}$ is set to zero for $t \in \{t| y_t \notin C_{{1:n}, T}\}$; otherwise, $g_t$ is set to one for $t \in \{t| y_t \in C_{{1:n}, T}\}$. 
To mitigate the length imbalance between the target terms and remaining words in the target sentence, we set $\alpha$ and $\beta$ to the value obtained by dividing their respective lengths from the total length.

\section{Experiments on the \texttt{HOLLY} benchmark}
\textcolor{black}{
In this section, we report the performance of our methodology when tested on the \texttt{HOLLY} benchmark.
In Section~\ref{sec:exp-homograph-disambiguation}, we evaluate the performance of our \emph{homograph disambiguation module} in determining the semantic appropriateness of a lexical constraint.
In Section~\ref{exp: LNMT}, we assess the performance of LNMT models using \textbf{\textit{positive examples}} from the \texttt{HOLLY} benchmark under conventional settings. 
Subsequently, we investigate the potential advantages that the homograph disambiguation module might bring when applied to the \textbf{\textit{negative examples}} from the \texttt{HOLLY} benchmark.
}
\subsection{Homograph Disambiguation} \label{sec:exp-homograph-disambiguation}
\subsubsection{Data}
Here, we present our dataset for training the homograph disambiguation module.
\textcolor{black}{Our training data was collected from the Korean dictionary\footnote{To release data, we collected the examples that are controlled by the appropriate license. CC BY-SA 2.0 KR.}} and \textcolor{black}{we manually inspected the quality of each sentence.}
In line with Fig.~\ref{fig:homograph disambiguation module}, each example consists of a triplet of example sentences containing a common homograph. 
\textcolor{black}{
Depending on the inter-sentential relationships between each input sentence, the homograph disambiguation module outputs a binary label: ``1'' is assigned if the homograph carries the same meaning in all sentences, and ``0'' if used differently in one example sentence.
}
The brief data statistics \textcolor{black}{of the training data} are reported in Table~\ref{tab:homograph-data-statistics}.
Note that any homographs are not allowed to be overlapped across train, validation, and test datasets.
\begin{table}[h!]
\centering
   \resizebox{1.0\columnwidth}{!}{%
\begin{tabular}{c|c|ccc}
\hline\hline
\multirow{2}{*}{} & \multirow{2}{*}{\makecell{\textbf{\# of words}\\\textbf{(homograph)}}} & \multicolumn{3}{l}{$\;\;\;\;\;\;\;\;$\textbf{\hfil\# of examples}}                                  \\ \cline{3-5} 
                  &                              & \multicolumn{1}{l|}{\textbf{Class 1}} & \multicolumn{1}{l|}{\textbf{Class 0}} & \textbf{Total} \\ \hline
\textbf{Train}             & 434   & \multicolumn{1}{l|}{\hfil13,128}   & \multicolumn{1}{l|}{\hfil35,708}   & \hfil48,836 \\ 
\textbf{Validation}        & 39                           & \multicolumn{1}{l|}{\hfil1,500}    & \multicolumn{1}{l|}{\hfil1,500}    &  \hfil3,000  \\ \hline
\end{tabular}}
\caption{Data statistics for the homograph disambiguation task.}
\vspace{-3mm}
\label{tab:homograph-data-statistics}
\end{table}

\textcolor{black}{At test time, we evaluated our model on the \texttt{HOLLY} benchmark.}
\textcolor{black}{Specifically, for each lexical constraint, two positive references (refer to Table~\ref{table: positive references}) and one of the four test example sentences ((a), (b), (c), or (d) in Table~\ref{table: evaluation data}) is given as a triplet.}
\subsubsection{Results}
We conducted experiments with two well-known variants of PLM trained on Korean corpora: \textit{klue/roberta-base}, and \textit{klue/roberta-large}.
\textcolor{black}{
Our homograph disambiguation module achieved a test accuracy of 88.7\%, and 92.3\% when using \textit{klue/roberta-base}, and \textit{klue/roberta-large}, respectively.}
\textcolor{black}{
In spite of the imbalanced data distribution shown in Table \ref{tab:homograph-data-statistics}, the values of precision and recall are balanced in both classes, as shown in Table \ref{tab:homograph-test-accuracy}.}
\begin{table}[h!]
\centering
   \resizebox{\columnwidth}{!}{%
\begin{tabular}{c|cc|c}
\hline\hline
$\;\;\;\;\;\;$\textbf{Class}$\;\;\;\;\;\;$ & $\;\;\;\;\;\;$\textbf{Precision}$\;\;\;\;\;\;$ & $\;\;\;\;\;\;$\textbf{Recall}$\;\;\;\;\;\;$ & $\;\;\;\;\;\;\;\;$\textbf{F1}$\;\;\;\;\;\;\;\;$    \\ \hline
1     & 0.924    & 0.933  & 0.929  \\  
0     & 0.933    & 0.923  & 0.928  \\ \hline
\end{tabular}}
\caption{\textcolor{black}{Test accuracy of homograph disambiguation module leveraging \textit{klue/roberta-large} on the \texttt{HOLLY} benchmark. The accuracy is reported in terms of precision, recall, and F1 on each class. }}
\vspace{-3mm}
\label{tab:homograph-test-accuracy}
\end{table}

\subsection{Lexically-constrained NMT}
\label{exp: LNMT}

\subsubsection{Training Data} \label{sec:filtering}

\textcolor{black}{
We used 1.83M sentence pairs from two publicly available Korean-English datasets as training corpora: IWSLT 17 training data and AI Hub parallel data.}\footnote{The AI HUB data can be found here: \url{https://www.aihub.or.kr/aihubdata/data/view.do?currMenu=115&topMenu=100&aihubDataSe=realm&dataSetSn=126}.}
We pre-tokenized the Korean corpora with Mecab and built a joint vocabulary for both languages by \textcolor{black}{learning a Byte
Pair Encoding~\cite{sennrich2016neural} model in sentencepiece~\cite{kudo2018sentencepiece} with 32K merge operations.}

\textcolor{black}{To simulate the unseen lexical constraints, we filtered out about 160K training sentence pairs with test lexical constraints on both sides when experimenting with the \texttt{HOLLY} benchmark.}
\textcolor{black}{
This filtering process
is crucial for examining how the models cope with any lexical constraints that users might introduce.
}
\subsubsection{Evaluation}
\paragraph{Metrics}
We evaluated the performance of our model in terms of BLEU\footnote{We measure the BLEU scores using sacreBLEU~\cite{post2018call} with the signature nrefs:1|case:mixed|eff:no|tok:13a|smooth:exp|version:2.0.} and copy success rate (CSR). CSR is a metric for investigating the ratio of imposed lexical constraints met in translations.
\textcolor{black}{For a statistical significance test, we use \texttt{compare-mt}~\cite{neubig2019compare} with $p=0.05$ and 1,000 bootstraps.}

\begin{table*}[h!]
\centering
   \resizebox{2\columnwidth}{!}{%
\begin{tabular}{c|c|c|c}
\hline\hline
\textbf{Test Case} & \textbf{Lexical Constraint} & \textbf{Test Example} & \textbf{Expected Target Term(s)}\\ \hline
\multirow{2}{*}{\textbf{Soft Matching}} & \multirow{2}{*}{\begin{CJK}{UTF8}{mj}\colorbox{lime}{\textcolor{blue}{\textbf{소화}}}\end{CJK} $\rightarrow$ digest}    & \multirow{4}{*}{\begin{tabular}[c]{@{}l@{}}Src. \begin{CJK}{UTF8}{mj}사람의 이는 음식물을 잘게 부숴 삼키기 좋게 하여 \colorbox{lime}{\textcolor{blue}{\textbf{소화}}}를 돕는 역할을 한다.\end{CJK}\\ Ref. The human teeth function to break down food items into comfortably swallowable \\pieces, helping \hl{digestion}.\end{tabular}} & \multirow{2}{*}{digest | digestion | digestive} \\
                        &                                             &                                                                                                                                                                                                                  &                                                 \\ \cline{1-2} \cline{4-4} 
\multirow{2}{*}{\textbf{Hard Matching}} & \multirow{2}{*}{\begin{CJK}{UTF8}{mj}\colorbox{lime}{\textcolor{blue}{\textbf{소화}}}\end{CJK} $\rightarrow$ \hl{digestion}} &     & \multirow{2}{*}{\hl{digestion}}                      \\
                        &                                             &                                                                                                                                                                                                                  &                                                 \\ \hline
\end{tabular}}
\caption{Two test scenarios are described with one of test example in the \texttt{HOLLY} benchmark. The source term can be pronounced as ``so-hwa''.}
\vspace{-2mm}
\label{tab: test scenarios}
\end{table*}

\paragraph{Test Scenarios}
\textcolor{black}{There were two important test cases, as shown in Table~\ref{tab: test scenarios}.
Given a source sentence, we can consider the \textbf{\emph{Soft Matching}} test case, which allows some morphological variations, as introduced in~\citet{Dinu}.
As illustrated in Table~\ref{tab: test scenarios}, since the Korean word \begin{CJK}{UTF8}{mj}\colorbox{lime}{\textcolor{blue}{\textbf{소화}}}\end{CJK} can be used in multiple different forms via inflection, any one of the expected candidates (digest, digestion, and digestive) presented in the translation is considered to be correct in terms of CSR.
}

\textcolor{black}{
We also have the \textbf{\emph{Hard Matching}}\footnote{This test case is suggested by previous work~\citep{LeCA, Song, Cdalign}.} test case where the exact target term (e.g., \hl{digestion}) presented in its reference has to be incorporated in the translation.
Note that, this cannot be tested on negative examples since the target terms in lexical constraints do not appear in their references.}

\paragraph{Baselines}
\begin{itemize}
    \setlength\itemsep{0em}
    \item Code-Switching (CS)~\cite{Song} replaces source terms \textcolor{black}{with} aligned target terms and learns to copy them via pointer network.
    \item LeCA~\cite{LeCA} modifies the source sentence as described in Table~\ref{tab:input-data}, and utilizes pointer network during training.
    \item Cdalign~\cite{Cdalign} proposes constrained decoding based on alignment.\footnote{We compare our model to the ATT-INPUT approach, which suffers from high time complexity but guarantees a high CSR.}
\end{itemize}

\pgfplotstableread{
method     TEST1     TEST2    
PLUMCOT    96.75     9.42
CS         88.96     7.47
Cdalign    89.94     7.47    
LeCA       89.94     7.14   
}\mytable

\begin {figure*}[ht]
\begin{tikzpicture} 
    \hspace*{4em}
    \vspace*{-3em}
    \begin{axis}[
    hide axis,
    xmin=-2.5,xmax=2.5,ymin=-2.5,ymax=2.5,
    axis lines=center,
    legend style={
            anchor=north,
            legend columns=-1,
            /tikz/every even column/.append style={column sep=0.5cm}
        },
    legend image code/.code={
        \draw[#1] (0cm,-0.2cm) rectangle (0.2cm,0.2cm);}
    ]
    \addlegendimage{mark=square*,color=blue, fill=blue!30},
    \addlegendentry{$\;$w/o correction};
    \addlegendimage{mark=square*,color=red, fill=red!30},
    \addlegendentry{$\;$w/ correction};
    \end{axis}

\end{tikzpicture}
\vspace*{-3em}
\newline
 \begin{tikzpicture}
    \begin{axis}  
    [  
        ybar, 
        axis on top,
        title={BLEU Score},
        bar width=0.6cm,
        enlargelimits=0.15,
        ymajorgrids, tick align=inside,
        symbolic x coords={PLUMCOT, CS, Cdalign, LeCA},  
        xtick=data,
        every node near coord/.append style={font=\footnotesize},
        nodes near coords={
            \pgfmathprintnumber[precision=1]{\pgfplotspointmeta}
           },
        nodes near coords align={vertical},
        ]  
    \addplot coordinates {(PLUMCOT, 16.49) (CS, 14.29) (Cdalign, 15.29) (LeCA, 15.71)};
    \addplot coordinates {(PLUMCOT, 18.30) (CS, 17.27) (Cdalign, 17.54) (LeCA, 17.73)}; 
    \end{axis}  
  \end{tikzpicture}  $\quad\:\quad\quad$
\begin{tikzpicture}
    \begin{groupplot}[
        group style={
            group size=1 by 2,
            vertical sep=0.1cm,
        },
        ybar,
        symbolic x coords={PLUMCOT, CS, Cdalign, LeCA},
      ]
      \nextgroupplot[
          bar width=0.6cm,
          ymajorgrids, tick align=inside,
          title={CSR (\%)},
          axis y line*=right,
          enlargelimits=0.15,
          xtick=\empty,
          ymin=80,
          ymax=100,
          axis x line*=top,
          nodes near coords={
          \pgfmathprintnumber[precision=1]{\pgfplotspointmeta}
         },
         height=2.435cm,
         every node near coord/.append style={font=\footnotesize},
         nodes near coords align={vertical}, 
        ]

      \begin{scope}

        \clip
          (axis cs: {[normalized]-1}, 80)
                      rectangle
          (axis cs: {[normalized] 8},  100);
        \addplot table[x=method, y=TEST1] {\mytable};
        \addplot table[x=method, y=TEST2] {\mytable};
      \end{scope}

      \nextgroupplot[
          bar width=0.6cm,
          enlargelimits=0.15,
          xticklabels={PLUMCOT, CS, Cdalign, LeCA},
          ymajorgrids, tick align=inside,
          xtick=data,
          ymax=20,
          axis y line*=right,
          axis x line*=bottom,
          every node near coord/.append style={font=\footnotesize},
          nodes near coords={
          \pgfmathprintnumber[precision=1]{\pgfplotspointmeta}
         },
         height=2.835cm,
         nodes near coords align={vertical},  
        ]

        \addplot table[x=method, y=TEST1] {\mytable};
        \addplot table[x=method, y=TEST2] {\mytable};
    \end{groupplot}
  \end{tikzpicture}
\vspace{-3mm}
\caption{Effect of homograph disambiguation tested on negative examples. \textcolor{black}{``w/ correction'' refers to the removal of semantically inappropriate lexical constraints determined by the homograph disambiguation module.} \textcolor{black}{CSR was evaluated on \emph{Soft Matching}.}}
\vspace{-3mm}
\label{fig:benefit-of-correction}
\end{figure*}
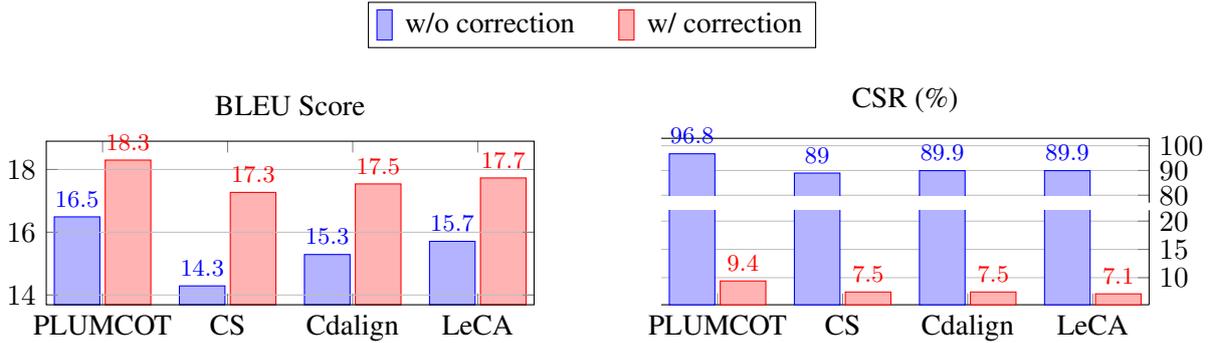
\subsubsection{Main Results}

\paragraph{Simulating Unseen Lexical Constraints}
Table~\ref{tab:vanilla-filtering-effect} shows the importance of simulating unseen lexical constraints.
When lexical constraints are exposed during training, the vanilla Transformer already achieves 66.67\% of CSR by mere memorization.
We observe that eliminating around 160K overlapping training examples results in a significant reduction of CSR (66.67\% $\rightarrow$ 11.97\%), indicating that we manage to simulate the conditions where lexical constraints are nothing short of unseen.
\begin{table}[h!]
\centering
   \resizebox{1\columnwidth}{!}{%
\begin{tabular}{ c | c | c | c | c }
\hline\hline
    \multirow{2}{*}{Method} & \multicolumn{2}{c|}{Soft Matching} & \multicolumn{2}{c}{Hard Matching} \\ \cline{2-5}
          & BLEU & CSR (\%) & BLEU & CSR (\%) \\
    \hline
    Trained w/ \emph{filtered} data  & 18.44 & 11.97 & 18.44 & 9.06 \\
    Trained w/ \emph{full} data & 21.65 & 66.67 & 21.65 & 58.90 \\
    \hline
\end{tabular}}
\caption{Performance of the vanilla Transformer on positive examples. Without filtering, lexical constraints can be memorized by the network during training.}
\vspace{-2mm}
\label{tab:vanilla-filtering-effect}
\end{table}

\paragraph{Results on Positive Examples}
The performances \textcolor{black}{of LNMT models} on positive examples\footnote{Recall that positive examples are bound together with semantically appropriate lexical constraints. Refer to (a) and (b) in Table~\ref{table: evaluation data} for details.} are compared in Table~\ref{tab:HOLLY-positive}.
It is shown that \texttt{PLUMCOT} outperforms all the baselines in both metrics by a large margin.
Since we simulate the unseen lexical constraints, the external information from the PLM contributes to the increase in BLEU.
Combined with the \textcolor{black}{supervision on a copying score}, \texttt{PLUMCOT} achieves the highest CSR.\footnote{\textcolor{black}{Instead of using the \texttt{HOLLY} benchmark, we also tested the performance of \texttt{PLUMCOT} in a test benchmark used in previous studies~\cite{Cdalign, LeCA} to compare its effectiveness, as shown in Appendix~\ref{appendix: Randomly Sampled Test Constraints}.}}
The overall BLEU scores of \emph{Hard Matching} are shown to be greater than \emph{Soft Matching} as the expected target terms drawn from \emph{reference translations} are given to the models.
\begin{table}[h!]
\centering
\resizebox{1\columnwidth}{!}{%
\begin{tabular}{ c | c | c | c | c }
\hline
\hline
    \multirow{2}{*}{Method} & \multicolumn{2}{c|}{Soft Matching} & \multicolumn{2}{c}{Hard Matching} \\ \cline{2-5}
          & $\;\;\;$BLEU$\;\;\;$ & $\;\;\;$CSR (\%)$\;\;\;$ & $\;\;\;$BLEU$\;\;\;$ & $\;\;\;$CSR (\%)$\;\;\;$ \\
    \hline
    CS & 18.52 & 93.20 & 20.21 & 92.56 \\
    LeCA & 19.33 & 94.17 & 20.53 & 93.85 \\
    Cdalign & 17.02 & 95.47 & 17.52 & 94.82 \\
    \hline
    \textbf{\texttt{PLUMCOT} (Ours)}   & $\;\;$\textbf{20.91}\textnormal{\textsuperscript{$\star$}} & \textbf{98.06} & $\;\;$\textbf{22.07}\textnormal{\textsuperscript{$\star$}} & \textbf{98.71} \\ 
 \hline
\end{tabular}}
\caption{\textcolor{black}{LNMT performances on positive examples.} All the models are trained with the \emph{filtered} data as stated in Section~\ref{sec:filtering}. \textcolor{black}{``$\star$'' demonstrates that our method achieves statistically significant performance over baselines on positive examples.}}
\vspace{-2mm}
\label{tab:HOLLY-positive}
\end{table}

\paragraph{Benefits of Homograph Disambiguation}
Here, we analyze beneficial effects of homograph disambiguation on LNMT.
Since lexical constraints in negative examples\footnote{Refer to examples (c) and (d) in Table~\ref{table: evaluation data}.} are semantically improper, the homograph disambiguation module determines whether they should be imposed or not. 
Corrections are made with its decisions; the corresponding lexical constraints are removed.
The effects of the \emph{correction} are shown in Fig.~\ref{fig:benefit-of-correction}.

We observed significant drops in CSR across all of the models, which is desirable since the lexical constraints are irrelevant to the context.
\textcolor{black}{
By removing inappropriate constraints, all the models achieve a consistent \textcolor{black}{and statistically significant} improvement in translation quality by a large margin.}


\subsubsection{Ablation Study}\label{sec:ablation-plm-gate}
\textcolor{black}{
We study the effect of each component of \texttt{PLUMCOT}, and the results are provided in Table~\ref{tab:ablation study}.
Compared to the \texttt{PLUMCOT} without supervision, the supervision on a copying score significantly improves CSR  (93.85\% vs. 98.06\%).
We find that leveraging rich contextual representation of PLM can further improve the translation quality (18.51 $\rightarrow$ 20.91).
The BLEU score of a model that combines only PLM without supervision on a copying score is lower than that of the \texttt{PLUMCOT} model.
This may simply be due to a higher BLEU score from better reflecting the target terms in the positive examples (an increase in CSR from 93.85\% to 98.06\%).
Combining the two components yields the best performance in both metrics. More ablations can be found in Appendix~\ref{appendix: Ablation}.
}
\begin{table}[h!]
\centering
\resizebox{1\columnwidth}{!}{%
\begin{tabular}{ c | c | c | c | c }
\hline
\hline
    \multirow{2}{*}{Method} & \multicolumn{2}{c|}{Soft Matching} & \multicolumn{2}{c}{Hard Matching} \\ \cline{2-5}
          & BLEU & CSR (\%) & BLEU & CSR (\%) \\
    \hline
    \textbf{PLUMCOT} & \textbf{20.91} & \textbf{98.06} & \textbf{22.07} & \textbf{98.71} \\
    $(-)$ PLM & 18.51 & 98.06 & 19.58 & 98.38 \\
    $(-)$ Supervision & 19.22 & 93.85 & 20.54 & 93.20 \\ 
 \hline
\end{tabular}}
\caption{Ablation studies performed on positive examples \textcolor{black}{\emph{w/o} correction. ``PLM'': integration of PLM. ``Supervision": supervised learning of a copying score.}}
\vspace{-3mm}
\label{tab:ablation study}
\end{table}

\begin{table*}[h!]
\centering
\resizebox{2\columnwidth}{!}{%
\begin{tabular}{ll}
\hline\hline
\multicolumn{2}{l}{\textbf{Positive Example (Lexical Constraint: \begin{CJK}{UTF8}{mj} \colorbox{lime}{\textbf{\textcolor{blue}{세제}}}\end{CJK} $\rightarrow$ \hl{tax system})}}                                                                                                                               \\ \hline
\multirow{2}{*}{\textbf{Source}}                 & \multirow{2}{*}{\begin{CJK}{UTF8}{mj}거래세를 줄이고 보유세를 강화하는 게 부동산 \colorbox{lime}{\textbf{\textcolor{blue}{세제}}}의 대원칙이지만 이를 적용하기도 어렵다.\end{CJK}}                                                                                                         \\
                                                 &                                                                                                                                                                            \\
\multirow{2}{*}{\textbf{Reference}}              & \multirow{2}{*}{Reducing transaction taxes and raising possession taxes are the core principles of the real estate \hl{\textbf{tax system}}, but it is challenging to get them applied.} \\
                                                 &                                                                                                                                                                            \\ \hline
\multirow{2}{*}{\textbf{Vanilla}}    & \multirow{2}{*}{Reducing transaction taxes and strengthening holding taxes are the grand principles of real estate \uwave{taxes}, but it is also difficult to apply them.$\;$$$\XSolidBrush$$}         \\
                                                 &                                                                                                                                                                            \\
\multirow{2}{*}{\textbf{LeCA}}                   & \multirow{2}{*}{Reducing transaction taxes and strengthening holding taxes are the main principles of real estate \uwave{tax}, but it is difficult to apply them.$\;$$$\XSolidBrush$$}               \\
                                                 &                                                                                                                                                                            \\ \hline
\multirow{2}{*}{\textbf{\texttt{PLUMCOT}}}                & \multirow{2}{*}{The main principle of the real estate \hl{\textbf{tax system}} is to reduce transaction taxes and strengthen holding taxes, but it is difficult to apply them.$\;$$$\CheckmarkBold$$}          \\
                                                 &                                                                                                                                                                            \\ \hline
                                                 &                                                                                                                                                                            \\ \hline\hline
\multicolumn{2}{l}{\textbf{Negative Example (Lexical Constraint: \begin{CJK}{UTF8}{mj}\colorbox{lime}{\textbf{\textcolor{red}{세제}}}\end{CJK} $\rightarrow$ \hl{tax system})}}                                                                                                                               \\ \hline
\multirow{2}{*}{\textbf{Source}}                 & \multirow{2}{*}{\begin{CJK}{UTF8}{mj}이번 행사 기간 동안 5만 원 이상 구입하시는 고객에게는 주방 \colorbox{lime}{\textbf{\textcolor{red}{세제}}}를 경품으로 드립니다.\end{CJK}}                                                                                                         \\
                                                 &                                                                                                                                                                            \\
\multirow{2}{*}{\textbf{Reference}}              & \multirow{2}{*}{For those who spend more than 50 thousand won for purchasing items during this event, kitchen \colorbox{yellow}{\textbf{detergents}} will be given as a gift.}                         \\
                                                 &                                                                                                                                                                            \\ \hline
\multirow{2}{*}{\textbf{\texttt{PLUMCOT} w/o correction}} & \multirow{2}{*}{The kitchen \hl{\textbf{tax system}} will be given as a prize to customers who purchase more than 50,000 won during this event.}                                         \\
                                                 &                                                                                                                                                                            \\
\multirow{2}{*}{\textbf{\texttt{PLUMCOT} w/ correction}}  & \multirow{2}{*}{Customers who purchase more than 50,000 won during this event will receive a gift of kitchen \colorbox{yellow}{\textbf{detergent}}.}                                                   \\
                                                 &                                                                                                                                                                            \\ \hline
\end{tabular}}
\caption{Example translations for positive and negative examples. The source term can be pronounced as ``se-je''.}
\vspace{-1mm}
\label{tab:generation_example}
\end{table*}
\subsection{Qualitative Analysis}
Table~\ref{tab:generation_example} provides translated examples.
Given a lexical constraint, \texttt{PLUMCOT} incorporates the target term correctly.
In a negative example, the meaning of \begin{CJK}{UTF8}{mj} \colorbox{lime}{\textbf{\textcolor{red}{세제}}}\end{CJK} is properly translated into \colorbox{yellow}{detergent} by \texttt{PLUMCOT} with \emph{correction}.\footnote{Note that the correction is made by homograph disambiguation.}
\textcolor{black}{We provide more examples in Table~\ref{tab:more generation_example}.}

\section{Related Work}
\subsection{Lexically-constrained NMT}
\textcolor{black}{
Recent work on LNMT broadly falls into two categories: \emph{decoding algorithms} and \emph{inline annotation}.
During beam search, decoding algorithms enforce target terms to appear in the output~\citep{GBS, CBS, chatterjee-etal-2017-guiding, hasler-etal-2018-neural}.
This approach ensures a high CSR, but the decoding speed is significantly degraded.
}
\textcolor{black}{
To alleviate this issue,~\citet{post-vilar-2018-fast} suggests a decoding algorithm with a complexity of $O(1)$ in the number of constraints.}
Another variation on decoding algorithms utilizes word alignments between source and target terms~\citep{Song-align, Cdalign}. 

\textcolor{black}{In \emph{inline annotation} studies, the model is trained to copy target terms via modification of training data.
Either a source term is replaced with the corresponding target term, or the target term is appended to the source sentence~\citep{Song, Dinu, LeCA}. 
}
\textcolor{black}{
Concurrently,~\citet{bergmanis2021facilitating, niehues2021continuous, xu2021rule} consider the morphological inflection of lexical constraints during the integration of target terms.
}
While these methods incur a slight computational cost and \textcolor{black}{provide better translation quality}, target terms are not guaranteed to appear~\cite{Cdalign, wang2022template}.
To better copy target terms in a source sentence, a pointer network~\cite{pointer, pointing} that uses attention weights to copy elements from a source sentence is introduced~\cite{gu2019pointer, Song, LeCA}.
In this work, we further enhance the copying mechanism of a pointer network via supervised learning of a copying score that achieves better performance in terms of BLEU and CSR.


\subsection{Homograph Issue in LNMT}
\textcolor{black}{\citet{michon2020integrating} points out the homograph issue in LNMT in an in-depth error analysis of their model.}
To the best of our knowledge, the homograph issue was explicitly addressed first in ~\citet{oz2021towards}.
In their work, given a source homographic term, the most frequent alignment is selected as its correct lexical constraint, while the other alignments are treated as negative terms that should be avoided in the translation.
However, low-frequency meanings are important for LNMT since it is not guaranteed that users always bring up generic terminology.

Different from their method, our homograph disambiguation module infers the meaning of lexical constraints and makes decisions to impose them or not.
Furthermore, we confirm that our method works equally well on ``unseen'' homographs.

\subsection{Integration of PLM with NMT}

Followed by the success of PLM, researchers attempted to distill the knowledge of PLM into NMT~\cite{Bert-fused, weng2020acquiring, BiBERT}. 
BERT-fused~\cite{Bert-fused} is one such method; it plugs the output of BERT into the encoder and decoder via multi-head attention.
\textcolor{black}{We borrowed the idea from BERT-fused, and for the first time, combined LNMT and PLM, which works well even in ``unseen'' lexical constraints by leveraging the rich contextual information of PLM.}

\section{Conclusions}
\textcolor{black}{
In this paper, we investigate two unexplored issues in LNMT and propose a new benchmark named \texttt{HOLLY}.
To address the homograph issue of the source terms, we built a homograph disambiguation module to infer the exact meaning of the source terms.
We confirm that our homograph disambiguation module alleviates mistranslation led by semantically inappropriate lexical constraints.}
\texttt{PLUMCOT} is also proposed to improve LNMT by using the rich information of PLM and ameliorating its copy mechanism via direct supervision of a copying score.
\textcolor{black}{
Experiments on our \texttt{HOLLY} benchmark show that \texttt{PLUMCOT} significantly outperforms existing baselines in terms of BLEU and CSR. 
}

\section{Limitations}  
\textcolor{black}{Our study includes some limitations that must be addressed. Some test examples might have wrong predictions made by the \emph{homograph disambiguation module}.}
Specifically, in positive examples where lexical constraints should be imposed, its errors result in wrong corrections (i.e., the elimination of necessary lexical constraints).
Table~\ref{tab:limitation positive} shows how these erroneous corrections affect the results.

\begin{table}[h!]
\centering
\resizebox{1\columnwidth}{!}{%
\begin{tabular}{ c | c | c | c | c }
\hline
\hline
    \multirow{2}{*}{Method} & \multicolumn{2}{c|}{w/o correction} & \multicolumn{2}{c}{w/ correction} \\ \cline{2-5}
          & $\;\;\;$BLEU$\;\;\;$ & $\;\;\;$CSR (\%)$\;\;\;$ & $\;\;\;$BLEU$\;\;\;$ & $\;\;\;$CSR (\%)$\;\;\;$ \\
    \hline
    CS & 18.52 & 93.20 & 18.88 & 85.44 \\
    LeCA & 19.33 & 94.17 & 19.24 & 88.03 \\
    Cdalign & 17.02 & 95.47 & 17.03 & 89.97 \\
    \hline
    \textbf{\texttt{PLUMCOT} (Ours)}   & \textbf{20.91} & \textbf{98.06} & \textbf{20.80} & \textbf{91.59} \\ 
 \hline
\end{tabular}}
\caption{Effect of homograph disambiguation tested on positive examples \textcolor{black}{on \textbf{\emph{Soft Matching}}.}}
\vspace{-3mm}
\label{tab:limitation positive}
\end{table}
We can observe an overall decline in CSR; however, it does not hurt the translation quality. 
\textcolor{black}{We verify that the differences in BLEU resulting from wrong corrections are not statistically significant for all the methods.}
Considering the gain achieved in negative examples, as seen in Fig.~\ref{fig:benefit-of-correction}, our proposed homograph disambiguation might serve as a useful starting point to address homographs in LNMT; however, there is still room for improvement.
Our current \emph{homograph disambiguation module} is designed as a stand-alone system outside the LNMT. However, building an end-to-end system can be beneficial, which can be addressed in future work. 
\textcolor{black}{\section*{Acknowledgments}
Authors would like to thank all Papago team members for the insightful discussions. Also, we sincerely appreciate the fruitful feedback from Won Ik Cho and CheolSu Kim. We thank the anonymous reviewers for their valuable suggestions for enhancing this work. This work was supported by Papago, NAVER Corp, the Institute of Information \& communications Technology Planning \& Evaluation (IITP) grant funded by the Korea government (MSIT) (No.2019-0-00075, Artificial Intelligence Graduate School Program (KAIST)), and the National Research Foundation of Korea (NRF) grant funded by the Korea government (MSIT) (No. NRF-2022R1A2B5B02001913).}

\bibliography{anthology,custom}
\bibliographystyle{acl_natbib}
\appendix

\section{\texttt{HOLLY} benchmark} \label{appendix: Source of Data}
We collected monolingual example sentences that contain one of the pre-specified homographs from the Korean dictionary. For each homograph, retrieved example sentences are classified into multiple groups according to their meanings.
We chose one group with the least frequent meaning and used its examples as positive references to determine a lexical constraint for the homograph.
Examples from the other groups are considered as negative references.
Eventually, six reference sentences were collected for each homograph; more specifically, four positive references and two negative references.

Setting aside two positive references for homograph disambiguation, as stated in Table~\ref{table: positive references}, we outsourced the translation of two positive and negative examples, as introduced in Table~\ref{table: evaluation data}. \textcolor{black}{For positive examples, professional translators were requested to translate source terms of lexical constraints into pre-defined target terms. We guide the translators to carefully translate negative examples by focusing on the exact meaning of lexical constraints. }

\section{Implementation details}
%
\subsection{Configuration of \texttt{PLUMCOT}}
We implemented \texttt{PLUMCOT} and all the models based on fairseq~\cite{fairseq}. We matched the embedding dimensions, the number of layers, and the number of attention heads of all models for a fair comparisons. \texttt{PLUMCOT} was trained from scratch and \textit{klue/roberta-large}~\cite{KLUE} was used for our PLM.\footnote{Please refer to~\ref{Appendix: Model Config} for more details.}

\subsection{Computational Cost}
All the experiments were conducted on a single A100 GPU. It takes about 84 hours to train \texttt{PLUMCOT} and 5 hours to train the homograph disambiguation module. The number of training / total parameters for \texttt{PLUMCOT} is 156M and 493M. The number of training / total parameters for the homograph disambiguation module is 6M and 343M.

\section{Ablation Studies}
\label{appendix: Ablation}
\subsection{Weights of the supervised learning of a copying score}
The results in Table~\ref{tab:lambda} were reported according to the different weights $\lambda$ of the supervised learning of the copying score in Eq.~\eqref{eqn: gate classification}. Based on experimental results, we were able to find a compromise where $\lambda$ is 0.2.
\begin{table}[h!]
\centering
\resizebox{1\columnwidth}{!}{%
\begin{tabular}{ c | c | c | c | c }
\hline
\hline
    \multirow{2}{*}{Method} & \multicolumn{2}{c|}{Soft Matching} & \multicolumn{2}{c}{Hard Matching} \\ \cline{2-5}
          & $\;\;\;$BLEU$\;\;\;$ & $\;\;\;$CSR (\%)$\;\;\;$ & $\;\;\;$BLEU$\;\;\;$ & $\;\;\;$CSR (\%)$\;\;\;$ \\
    \hline
    PLUMCOT $(\lambda=1)$ & 19.00 & 99.68 & 20.49 & 99.35 \\
    PLUMCOT $(\lambda=0.2)$ & 20.91 & 98.06 & 22.07 & 98.71 \\ 
 \hline
\end{tabular}}
\caption{Results of BLEU and CSR according to different $\lambda.$}
\label{tab:lambda}
\end{table}

\subsection{Number of example sentences}
\label{appendix: Number of example sentences}

We experimented with a varying number of example sentences.
As we use more example sentences, the information from the inter-sentential relationships becomes richer, eventually improving homograph disambiguation performance.
Experiments with $n=1, 2,$ and $3$ show an accuracy of $91.33$\%, $92.33$\%, and $92.67$\%, respectively.
Although an experiment with $n=3$ provides the best accuracy, collecting positive references can sometimes be burdensome to users.
Therefore, we conclude that $n$ should be decided by considering its trade-off.

\subsection{Randomly Sampled Test Constraints}
\label{appendix: Randomly Sampled Test Constraints}
Different from our \texttt{HOLLY} benchmark, at test time, lexical constraints were \emph{randomly} sampled from the alignments in each sentence pair in previous studies ~\citep{Dinu, Song, LeCA, Cdalign, Vecconst}.
Ten different test sets were built based on ten randomly sampled sets of lexical constraints, as described in~\citet{Cdalign, LeCA}.
Test statistics are reported in Table~\ref{tab:CS-test}.
\textcolor{black}{It is shown that \texttt{PLUMCOT} achieves the highest BLEU. The CSR is slightly lower than Cdalign, indicating that the gain for ``seen'' constraints is insignificant.\footnote{Note that the models are trained with \emph{full} data as we cannot remove training examples that overlap with random test lexical constraints in advance.}
}
\begin{table}[h!]
\centering
\resizebox{1\columnwidth}{!}{%
\begin{tabular}{ c | c | c | c | c }
\hline
\hline
    \multirow{2}{*}{Method} & \multicolumn{2}{c|}{BLEU} & \multicolumn{2}{c}{CSR (\%)} \\ \cline{2-5}
          & $\;\;$Average$\;\;$ & $\;\;$STDEV$\;\;$ & $\;\;$Average$\;\;$ & $\;\;$STDEV$\;\;$ \\
    \hline
    $\:\:\:$Vanilla$\:\:\:$ & 19.14 & 0.00 & 81.67 & 0.00 \\
    CS & 20.95 & 0.13 & 94.66 & 0.00 \\
    LeCA & 22.10 & 0.05 & 96.33 & 0.00 \\
    Cdalign & 21.45 & 0.07 & \textbf{98.03} & 0.00 \\
    \hline
    PLUMCOT $(\lambda=0.2)$& \textbf{22.50} & 0.07 & 97.84 & 0.00 \\ 
 \hline
\end{tabular}}
\caption{Results on randomly sampled test lexical constraints. Statistics are drawn from five randomly constructed test datasets.}
\label{tab:CS-test}
\end{table}

\section{Equation Details}
\label{Appendix: MHA}
Let $Q$, $K$, and $V$ be the query, key, and value in~\cite{Transformer}, respectively. Then $\text{MHA}$ in Eq.~\eqref{eqn: Body MHA} and Eq.~\eqref{Eqn: Decoder multihead} can be calculated as
\begin{equation}\label{eqn: MHA}
\begin{aligned}
    \text{MHA}(Q,K,V)&= 
     \text{Cat}_{h=1}^H[\text{head}_1, \cdots, \text{head}_n]W_o,
     \\
    \text{head}_{i} &= 
     \text{Attn}( QW_{i}^{Q}, KW_{i}^{K}, VW_{i}^{V}),
     \\
     \text{Attn}(q,k,v) &= \text{softmax}(\dfrac{qk^T}{\sqrt{d_k}})v, 
\end{aligned}
\end{equation}
where the projection matrices are parameters $ W^{O} \in \mathbb{R}^{Hd_{v} \times \: d_{\text{model}}}$, 
$ W_{i}^{Q}, \: W_{i}^{K}, \: W_{i}^{V} \in \mathbb{R}^{d_{\text{model}} \times d_{k}}$ for $\text{MHA}$.
In this paper, we employ $d_{\text{model}}=768$, $H=12$, $d_k=d_v=d_{\text{model}} / H = 64$. Note that all baselines and our model \texttt{PLUMCOT} use the same number of heads and the same projection matrices size.
We use additional multi-head attention, $\text{MHA}_{B}$, which only differs in projection matrices size where
$ W_{i}^{Q}, \: W_{i}^{K}, \: W_{i}^{V} \in \mathbb{R}^{d_{\text{PLM}}  \times d_{k}}  \;$ for $\text{MHA}_{B}$.\footnote{Here, we utilize \textit{klue/roberta-large}, a RoBERTa-based PLM trained on Korean corpus. The size of $d_{\text{PLM}}$ is 1024.}

\begin{table*}[h!]
\centering
\resizebox{2\columnwidth}{!}{%
\begin{tabular}{ll}
\hline\hline
\multicolumn{2}{l}{\textbf{Positive Example (Lexical Constraint: \begin{CJK}{UTF8}{mj} \colorbox{lime}{\textbf{\textcolor{blue}{내성} }}(``nae-sung'')\end{CJK} $\rightarrow$ \hl{resistance})}}                                                                                                                               \\ \hline
\multirow{2}{*}{\textbf{Source}}                 & \multirow{2}{*}{\begin{CJK}{UTF8}{mj}항생제에 \colorbox{lime}{\textbf{\textcolor{blue}{내성}}}이 있는 새로운 종류의 병원균이 등장해서 국민의 건강을 위협하고 있다.\end{CJK}}                                                                                                         \\
                                                 &                                                                                                                                                                            \\
\multirow{2}{*}{\textbf{Reference}}              & \multirow{2}{*}{New types of pathogens with \hl{\textbf{resistance}} to antibiotics have emerged, threatening public health.} \\
                                                 &                                                                                                                                                                            \\ \hline
\multirow{2}{*}{\textbf{Vanilla}}    & \multirow{2}{*}{A new type of pathogens that are \uwave{tolerant} of antibiotics have emerged, threatening the health of the people.$\;$$$\XSolidBrush$$}         \\
                                                 &                                                                                                                                                                            \\
\multirow{2}{*}{\textbf{LeCA}}                   & \multirow{2}{*}{A new type of pathogen that is \uwave{tolerant} of antibiotics has emerged, threatening the health of the people.$\;$$$\XSolidBrush$$}               \\
                                                 &                                                                                                                                                                            \\ \hline
\multirow{2}{*}{\textbf{\texttt{PLUMCOT}}}                & \multirow{2}{*}{A new type of pathogen that has \hl{\textbf{resistance}} to antibiotics has emerged, threatening the health of the people.$\;$$$\CheckmarkBold$$}          \\
                                                 &                                                                                                                                                                            \\ \hline
                                                 &                                                                                                                                                                            \\ \hline\hline
\multicolumn{2}{l}{\textbf{Negative Example (Lexical Constraint: \begin{CJK}{UTF8}{mj}\colorbox{lime}{\textbf{\textcolor{red}{내성}}}(``nae-sung'')\end{CJK} $\rightarrow$ \hl{resistance})}}                                                                                                                               \\ \hline
\multirow{2}{*}{\textbf{Source}}                 & \multirow{2}{*}{\begin{CJK}{UTF8}{mj}그가 말수가 적은 것은 \colorbox{lime}{\textbf{\textcolor{red}{내성}}}적인 성격에서 연유한다.\end{CJK}}                                                                                                         \\
                                                 &                                                                                                                                                                            \\
\multirow{2}{*}{\textbf{Reference}}              & \multirow{2}{*}{His being quiet is because of his \colorbox{yellow}{\textbf{introverted}} personality.}                         \\
                                                 &                                                                                                                                                                            \\ \hline
\multirow{2}{*}{\textbf{\texttt{PLUMCOT} w/o correction}} & \multirow{2}{*}{His low words are based on his \hl{\textbf{resistance}} to introverts.}                                         \\
                                                 &                                                                                                                                                                            \\
\multirow{2}{*}{\textbf{\texttt{PLUMCOT} w/ correction}}  & \multirow{2}{*}{His low-level words are related to his \colorbox{yellow}{\textbf{introverted}} personality.}                                                   \\
                                                 &                                                                                                                                                                            \\ \hline
                                                 \\
                                                 \\
\hline\hline
\multicolumn{2}{l}{\textbf{Positive Example (Lexical Constraint: \begin{CJK}{UTF8}{mj} \colorbox{lime}{\textbf{\textcolor{blue}{사유}}}(``sa-yoo'')\end{CJK} $\rightarrow$ \hl{reason})}}                                                                                                                               \\ \hline
\multirow{2}{*}{\textbf{Source}}                 & \multirow{2}{*}{\begin{CJK}{UTF8}{mj}회사 측은 계약 당사자 간 계약의 절차성을 \colorbox{lime}{\textbf{\textcolor{blue}{사유}}}로 계약 무효를 결정했다고 설명했다.\end{CJK}}                                                                                                         \\
                                                 &                                                                                                                                                                            \\
\multirow{2}{*}{\textbf{Reference}}              & \multirow{2}{*}{The company explained that the contract cancellation was decided because of the \hl{\textbf{reason}} relevant to contract procedures between the contract parties.} \\
                                                 &                                                                                                                                                                            \\ \hline
\multirow{2}{*}{\textbf{Vanilla}}    & \multirow{2}{*}{The company explained that it decided to nullify the contract because of the procedurality of the contract between the parties.$\;$$$\XSolidBrush$$}         \\
                                                 &                                                                                                                                                                            \\
\multirow{2}{*}{\textbf{LeCA}}                   & \multirow{2}{*}{The company explained that it decided to nullify the contract because of the procedurality of the contract between the parties.$\;$$$\XSolidBrush$$}               \\
                                                 &                                                                                                                                                                            \\ \hline
\multirow{2}{*}{\textbf{\texttt{PLUMCOT}}}                & \multirow{2}{*}{The company explained that it decided to nullify the contract on the \hl{\textbf{reason}} of the procedure of the contract between the parties.$\;$$$\CheckmarkBold$$}          \\
                                                 &                                                                                                                                                                            \\ \hline
                                                 &                                                                                                                                                                            \\ \hline\hline
\multicolumn{2}{l}{\textbf{Negative Example (Lexical Constraint: \begin{CJK}{UTF8}{mj}\colorbox{lime}{\textbf{\textcolor{red}{사유}}}(``sa-yoo'')\end{CJK} $\rightarrow$ \hl{reason})}}                                                                                                                               \\ \hline
\multirow{2}{*}{\textbf{Source}}                 & \multirow{2}{*}{\begin{CJK}{UTF8}{mj}자본주의 국가에서 \colorbox{lime}{\textbf{\textcolor{red}{사유}}} 재산은 소유자의 의사에 따라 처분할 수 있다.\end{CJK}}                                                                                                         \\
                                                 &                                                                                                                                                                            \\
\multirow{2}{*}{\textbf{Reference}}              & \multirow{2}{*}{In capitalist countries, \colorbox{yellow}{\textbf{private}} assets can be disposed of according to the will of their owners.}                         \\
                                                 &                                                                                                                                                                            \\ \hline
\multirow{2}{*}{\textbf{\texttt{PLUMCOT} w/o correction}} & \multirow{2}{*}{In capitalist countries, private property can be disposed of according to the \hl{\textbf{reason}} of the owner.}                                         \\
                                                 &                                                                                                                                                                            \\
\multirow{2}{*}{\textbf{\texttt{PLUMCOT} w/ correction}}  & \multirow{2}{*}{In capitalist countries, \colorbox{yellow}{\textbf{private}} property can be disposed of according to the owner's will.}                                                   \\
                                                 &                                                                                                                                                                            \\ \hline
                                                 
\end{tabular}
}
\caption{More translations for positive and negative examples.}
\label{tab:more generation_example}
\end{table*}

\begin{table*}
\centering
\caption{Hyperparameters and model configuration of \texttt{PLUMCOT}.}
\begin{tabular}{lc} 
\hline
NMT & \multicolumn{1}{l}{~ ~ ~ ~ ~  ~ Transformer}  \\ 
\hline
~ ~ ~encoder layers                                                                          & 6                                \\
~ ~ ~encoder embed dim                                                                       & 768                              \\
~ ~ ~encoder feed-forward dim                                                                & 3072                             \\
~ ~ ~encoder attention heads                                                                 & 12                               \\
~ ~ ~decoder layers                                                                          & 6                                \\
~ ~ ~decoder embed dim                                                                       & 768                              \\
~ ~ ~decoder feed-forward dim                                                                & 3072                             \\
~ ~ ~decoder attention heads                                                                 & 12                               \\
~ ~ ~positional encodings                                                                    & Sinusoidal                       \\
\begin{tabular}[c]{@{}l@{}}~ ~ ~max source positions\\\end{tabular}                          & 1024                             \\
~ ~ ~max target positions                                                                    & 1024                             \\
~ ~ ~segment embeddings                                                                      & True                             \\
~ ~ ~dropout                                                                                 & 0.3                              \\
                                                                                             &                                  \\ 
\hline
PLM                                                                                          & klue/roberta-large               \\ 
\hline
~ ~ ~encoder layers                                                                          & 24                               \\
~ ~ ~encoder embed dim                                                                       & 1024                             \\
~ ~ ~encoder feed-forward dim                                                                & 4096                             \\
~ ~ ~encoder attention heads                                                                 & 16                               \\
~ ~ ~positional encodings                                                                    & learned positional encodings     \\
~ ~ ~max source positions                                                                    & 514                              \\
~ ~ ~max target positions                                                                    & 514                              \\
~ ~ ~segment embeddings                                                                      & True                             \\
                                                                                             &                                  \\ 
\hline
Hyperparameter                                                                               & Value                            \\ 
\hline
~ ~ ~optimizer                                                                               & Adamw                            \\
~ ~ ~ ~ ~ ~ ~$\beta_1,\beta_2$ & (0.9, 0.98)                      \\
~ ~ ~ ~ ~ ~ ~weight decay                                                                            & 0.0                              \\
~ ~ ~max updates                                                                             & 130k                             \\
~ ~ ~learning rate                                                                           & 0.0005                           \\
~ ~ ~learning rate warmup                                                                    & 4000 steps                       \\
~ ~ ~warmup init learning rate                                                               & 1e-7                             \\
~ ~ ~lr scheduler                                                                            & inverse sqrt                     \\
~ ~ ~max tokens                                                                              & 4000                             \\
~ ~ ~update frequency                                                                        & 8                                \\
~ ~ ~clip grad norm                                                                          & 1.0                             
\end{tabular}\label{Appendix: Model Config}
\end{table*}

\section{Input data augmentation}
\label{appendix: input data augmentation}
\textcolor{black}{
As illustrated in Table~\ref{tab:input-data}, we modify a source sentence $X$ as $\hat{X}$ by appending <sep> tokens followed by target terms.
Since lexical constraints are domain-specific or user-provided terminology, we exclude the top 1,500 frequent words from a $32\text{K}$ joint dictionary.
In our training, we randomly sample at most 3 target terms from the target sentence. 
For each sentence, from 0 to 3 target terms are sampled following the distribution [0.3, 0.2, 0.25, 0.25].
}

\section{\textcolor{black}{Integration of PLM in Decoder}}
\label{Appendix: Integration of PLM in Decoder}
Here, we follow the same notations in Section~\ref{body: PLM}. Let $S^l$ denote the decoder output at $l^{th}$ layer. $s_{t}^{l}$ is the $t$-th element of $S^{l}$, and $S_{1:t}^l$ denotes $t$ number of elements from $s_1^l$ to $s_t^l$, masking elements from $t+1$ to the end. The output of each layer of the decoder can be calculated as
\begin{equation}
\label{Eqn: Decoder multihead}
\begin{aligned}
    \hat{s}_{t}^{l} &= \text{LN}\big(\text{MHA}(s_t^{l-1}, \, S_{1:t}^{l-1}, \, S_{1:t}^{l-1}) \big) +  s_t^{l-1} ,\\
    \tilde{s}_{t}^{l} &= 
    \frac{1}{2} \Big(\text{MHA}(\hat{s}_t^{l}, \, H^{L}, \, H^{L} ) \\
    &+ \text{MHA}_{\text{B}} (\hat{s}_t^{l}, \, B, \, B ) \Big)  + \hat{s}_t^{l}
    ,\\
    s_t^{l} &= \text{LN}\big(\text{FFN}\big(\text{LN}(\tilde{s}_{t}^{l})\big) + \tilde{s}_{t}^{l} \big).
\end{aligned}
\end{equation}

\section{\textcolor{black}{Pointer Network}}
\label{appendix: Pointer Network}
We use the same notations in Section~\ref{sec: PLUMCOT} and Appendix~\ref{Appendix: MHA} and~\ref{Appendix: Integration of PLM in Decoder}.
Let $|\hat{X}|$ denotes the length of the modified source sentence $\hat{X}$, and $(\alpha_{t,1}, \alpha_{t,2}, \cdots, \alpha_{t,|\hat{X}|})$ denotes the averaged attention weight of $\text{MHA}(\hat{s}_t^{L}, \, H^{L}, \, H^{L} )$ over the multiheads in Eq.~\eqref{Eqn: Decoder multihead}. Then our copying score $g_t^\text{copy}$ can be calculated as 
\begin{equation}
\begin{aligned}
    g_t^\text{copy} &=\sigma(W_g[c_t; s_t^L] + b_g), \\
    c_t &= \displaystyle\sum_{i=1}^{|\hat{X}|} \alpha_{t,i} \times h_i^L,
\end{aligned}
\end{equation}
where $c_t$ and $s_t^L$ are concatenated, $W_g$ and $b_g$ are the weight matrix and bias vector, and $\sigma(\cdot)$ is the sigmoid function.


\end{document}